\documentclass{article}
\usepackage{linguex}
\usepackage[english]{babel}
\usepackage[utf8]{inputenc}
\usepackage{johd}
\usepackage{booktabs}  
\usepackage{amsfonts} 
\usepackage[figuresright]{rotating}
\usepackage{multirow}

\title{Completing Networks by Learning Local Connection Patterns}

\author{Zhang Zhang$^{ab}$, Ruyi Tao.$^{ab}$$^{\dag}$, Yongzai Tao$^{c}$$^{\dag}$, Mingze Qi$^{d}$, Jiang Zhang$^{ab}$$^{*}$ \\
        \small $^{a}$School of Systems Science, Beijing Normal University \\
        \small $^{b}$Swarma Research, Beijing, China \\
        \small $^{c}$College of Computer Science and Technology, Zhejiang University \\
        \small $^{d}$College of Science, National University of Defense Technology \\\\
        \small $^{\dag}$Those author contribute equally\\
        \small $^{*}$Corresponding author: Jiang Zhang; \tt{zhangjiang@bnu.edu.cn}
        
}

\date{} 

\begin{document}

\maketitle

\begin{abstract} 
\noindent Network completion is a harder problem than link prediction because it does not only try to infer missing links but also nodes. Different methods have been proposed to solve this problem, but few of them employed structural information - the similarity of local connection patterns. In this paper, we propose a model named C-GIN to capture the local structural patterns from the observed part of a network based on the Graph Auto-Encoder framework equipped with Graph Isomorphism Network model and generalize these patterns to complete the whole graph. Experiments and analysis on synthetic and real-world networks from different domains show that competitive performance can be achieved by C-GIN with less information being needed, and higher accuracy compared with baseline prediction models in most cases can be obtained. We further proposed  a metric "Reachable Clustering Coefficient(CC)" based on network structure. And experiments show that our model perform better on a network with higher Reachable CC.
\end{abstract}

\noindent\\


\section{Introduction}

Networks build underlying structures in many systems and play important roles in science and daily lives \cite{marsden1990network,newman2012communities}. Thus one way to understand these complex systems is to study the property of the underlying networks. However the complete structure of a network usually can not be observed due to measurement errors, privacy concern and other reasons \cite{cimini2015systemic,kossinets2006effects,anand2018missing}. For example, we can gather online social network easily, but some offline nodes may have huge influence in some times but their information is hard to be gathered by these offline data. When facing the incomplete network data we need some methods to infer the missing information.

Inferring the complete network structure based on partially observable information has been a fundamental problem in network science and can benefit lots of downstream tasks and applications. Such as node classification \cite{kipf2016semi,bhagat2011node,rong2019dropedge}, graph classification \cite{cai2018comprehensive,bunke2008graph}, graph dynamics learning \cite{sanchez2020learning,wang2020pm2}, etc. In most of the network structure inferring works, researchers focus on a task where all nodes of a network can be observed but some edges are missing. This is called link prediction task \cite{lu2011link,wang2015link,lichtenwalter2010new}. One generally used idea to accomplish this task is by using structural information such as common neighbors \cite{zhou2009predicting,liu2011link} and mutual information \cite{tan2014link}. Another way to solve this problem is to use graph neural networks to learn which nodes should be connected by utilizing node features  \cite{kipf2016variational,zhang2018link} or structure patterns \cite{zhang2018link,feng2020link}. Besides Link prediction task, there is another task called network completion, which is harder because less information can be observed but same objective is remained.

In the task of network completion, we also need to infer the missing links between nodes but in which some nodes are also missing, and only the number of the nodes is known. This task is more difficult than link prediction because of the existence of ``naked nodes'' on the partial networks. ``Naked nodes'' refers to the nodes without any link which is isolated from all other nodes. There are no ``naked nodes'' in the link prediction task because all nodes will at least be connected to more than one neighbor. Thus, for a node with missing links, the information of its neighborhood can be used to infer its features, e.g., the methods based on common neighbors. However, in the network completion task, we cannot use any information of neighbor nodes for the ``naked nodes'' . Actually, the network completion task has similarity with the discovery of co-founders in causal inference \cite{pearl2010causal} which has wide applications because a missing node is an unobservable variable which however can take effects on observable variables \cite{chen2022inferring}. 

Some previous works have tried to tackle the problem of network completion. For example, G-GCN model \cite{xu2019generative} view the network completion problem as a network growth inference problem, and learn the network growth patterns from the observable part to infer the unobserved part. However, not all networks can be regarded as growing networks, for example, a trade network with countries as nodes is not suitable because on the one hand, the number of nodes is limited by the number of countries, on the other hand, the network may shrink, because there are cases where countries withdraw from trade relations and the nodes in the network disappear. The gene regulatory network is also not suitable to be regarded as a growing network. Of course, in the long evolutionary process, new genes are constantly generated \cite{betran2002expansion,long2003origin}. From this perspective, the gene regulatory network is a growing network. But on the time scale of a single species (which can be seen as a time slice in the evolutionary process), gene regulatory networks are usually considered to be invariant, and we usually do not study the growth properties of gene regulatory networks in individual species. Wei's model \cite{wei2021unifying} uses the attributes of nodes to initialize the estimated network topology, and then complete the network by refining the structure according to node attributes, labels and distances. Although both of the above works can achieve good performance on network completion, they all utilized the information of node attributes of the unobserved nodes, which reduces the difficulty of the problem. However, node attributes are always hard to be obtained in many real cases, which limits the applications of the above methods. 

The second method is by using the shared node dynamics to complete the network. For example, GGN model \cite{chen2022inferring} infers the information of the unobserved nodes and complete the entire network by utilizing time series data of observed nodes. By training a graph neural network to capture the node dynamics in common, GGN can infer the missing links of nodes including the unobserved ones. This method can complete the network with a high accuracy, but a large number of time-series data is needed. 

DeepNC \cite{tran2020deepnc} tackles the problem by training a GraphRNN to learn transferable connection patterns from a large number of 'similar graphs' to predict the unknown parts of a single incomplete graph, and achieves high performances. However, this is a totally different task as ours, and a huge amount of similar networks are hard to be found in most real cases, and only a single incomplete network can be used.

So far, to the best of our knowledge, the KronEM model \cite{kim2011network} is the only method to solve the single network completion problem without node information exactly as we defined. That is, neither time series data, node features nor a large number of similar networks is used. Compared to other methods above, KronEM uses minimal information and therefore has wide application scenarios. The basic idea of their work is that self similarity has been a basic property of many real world networks, so by observing the part of the network, they can infer a kernel to represent the relationship between the part and the whole of the network. Although network self-similarity is common in the real world, not all networks are self-similar \cite{song2005self}. In addition, the parameters of the model are too few to describe complex structures of a network precisely.

In this work we present a new model named Completion Graph Isomorphism Network(C-GIN) to solve the network completion problem without node information. The basic idea of C-GIN model is that different parts of one network follow some similar local connection patterns. We use a Graph Auto-Encoder to learn the patterns from observed part of the adjacency matrix and generalize them to unobserved parts. C-GIN model has a compact architecture and overcomes the shortcomings of the models we introduced above, the G-GCN model \cite{xu2019generative} and GGN model \cite{chen2022inferring} essentially rely on node information to complete the network(node feature for G-GCN and time-series data for GGN), and this information is not available in many cases. C-GIN model only takes the observable part of the adjacency matrix as input, and does not need node feature information, which makes C-GIN model have a wider range of application scenarios than these two models. Both DeepNC model \cite{tran2020deepnc} and C-GIN model complement the network through a structural pattern perspective. But the DeepNC model uses a large number of similar small networks(such as protein networks) with dozens of nodes, to learn the network structure pattern. C-GIN model uses the known local learned network structure pattern of a large network (more than hundreds of nodes). This makes C-GIN method more suitable for network completion task when some nodes in a network are missing. Compared to KronEM \cite{kim2011network}, C-GIN model is more powerful to represent more complex local network structures because graph neural networks always have more parameters. Experiments show that C-GIN model can complete networks with better performances.

\section{Related Works}
\subsection{Graph Auto-Encoder}
Graph Auto-Encoder \cite{wang2016auto,kipf2016variational}, which aims to encode a graph into a set of vectors in a continuous space to represent the structural information sufficiently with a trainable way \cite{li2021graph,do2020graph,li2019structure}, has obtained wide applications. Usually, one graph auto-encoder consists of two parts, one is the encoder which takes adjacency matrix $A_{n \times n}$(where $A_{i,j} = 1$ if node $v_i$ connects to node $v_j$ and $A_{i,j} = 0$ otherwise) and node attribute $X \in \mathbb{R}^{N \times k}$ as its input and turn it into a matrix $H \in \mathbb{R}^{N \times d}$ as the output, where $k$ represents the original feature dimension of the data, and $d$ is used as a hyperparameter to represent the hidden feature dimension of the data. Another part is the decoder which takes the output of the encoder, the vectors of the nodes $H$ as its input, and outputs a set of edge probabilities $P \in [0,1]^{N \times \frac{N}{2}}$. The objective of the decoder is to reconstruct the original adjacency matrix as accurate as possible. So for the parameter $\theta$ the loss function can be the cross entropy function as follows:

\begin{equation}
\mathcal{L}(\theta) = -\displaystyle\sum_{i=0}^{N-2} \sum_{j=i+1}^{N} A_{i,j}\log(P_{i,j}) + (1-A_{i,j})\log(1-P_{i,j}).
\end{equation}

Usually we only parameterize the encoder part, and leave the decoder as simple operations such as Inner Product or Distance calculation. In this way, we can get the embedding vectors of nodes from the encoder, and get the probability of connection between any two nodes through the decoder.

\subsection{Graph Isomorphism Network}
A lot of Graph Neural Networks(GNN) models can be used as the candidate of the encoder. In this work we choose Graph Isomorphism Network(GIN) \cite{xu2018powerful}. GIN is a powerful GNN model with a simple architecture but can generates more expressive node embeddings than other GNN models such as GCN \cite{kipf2016semi} and GAT \cite{2017Graph}. Besides its powerful representability, GIN model can help us to understand why the GNN models can achieve good results and what is the theoretical upper bound of the representability of GNNs. What's more, the upper bound can be achieved by GIN via a reasonable aggregate function as follows:

\begin{equation}
h_v^{(k)} = MLP^{k}((1+\epsilon)h_v^{(k-1)} + \sum_{\mu \in  \mathcal{N}(v)}h_u^{(k-1)}).
\end{equation}

In this function $\epsilon$ is a learnable parameter. It means that to obtain an updated node representation vector, we need to add all the current embedding vectors of itself and its neighbors together and to feed it into an MLP.

\subsection{G-GCN model}

G-GCN is a graph convolutional network model for growing graphs \cite{xu2019generative}. It is mainly used to solve the problem of predicting the links of new nodes in the process of graph growth, such as the cold start problem in the recommendation system. We can also regard it as a network completion problem, that is, some information of the network is observed (including node attributions and structure), but the structure of the remaining network is unknown. Assuming that the connections of the entire network follow the same pattern, according to knowing the connection patterns of observed part of the network, then it is possible to infer the connection of the unknown part according to the connection patterns of the observed part of the network.

During the training process, G-GCN randomly connects the unknown parts, and uses the GCN modules to go through two steps: firstly,  generating $\mu$ and $\sigma$ as the distribution of nodes, and sampling node representation from them; secondly, generating an adjacency matrix from these representation. The parameters in the entire model are optimized by Kullback-Leibler divergence, regularization constraints on distribution, and reconstruction accuracy on the generated adjacency matrix.

The model performs well in growing networks, such as citation networks. However, many networks don't increase node over time, such as gene networks, trade networks, etc., and the performance of the model may not good at this situation.

\subsection{KronEM}

KronEM model \cite{kim2011network} is proposed to solve the network completion problem for the networks that have self-similar structures. The model generates a complete network by doing kronecker products based on a kernel matrix recursively, and optimizes the parameters within the kernel by using the expectation-maximization approach and a scalable, metropolized Gibbs sampling method. Although the competitive performance can be obtained on many large networks, shortcomings can not be ignored. Firstly, the basic assumption that the network should be self-similar can not be satisfied by many real networks \cite{gallos2007review}. Second, this model uses too few parameters to represent the rich fine structures of real networks. All these limit its applications in real-world.

\section{The Network Completion Problem}
In this section we will introduce the formal definition of the network completion problem, and then we will propose C-GIN model based on the graph auto-encoder framework to solve the problem.

\subsection{Problem Definition}
\begin{figure}[!ht]
\centering
\includegraphics[scale=0.4]{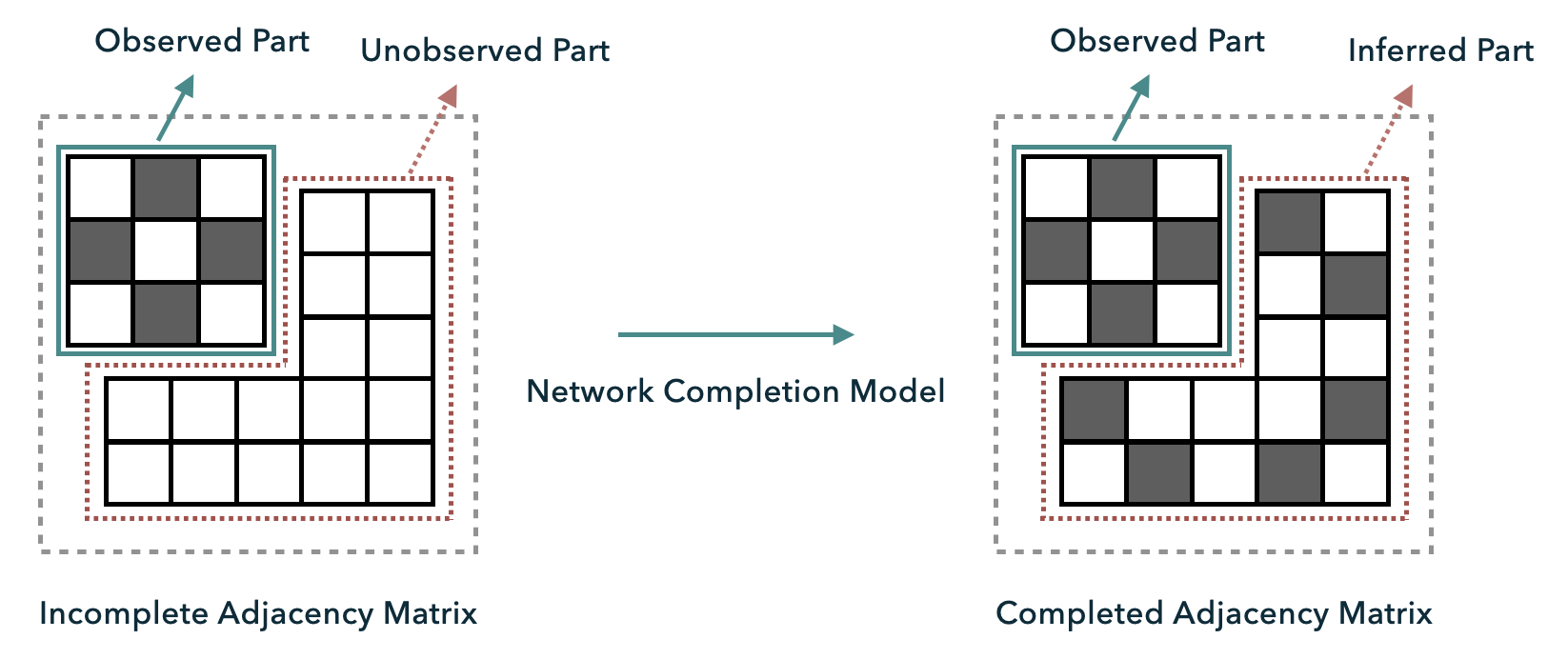}
\caption{\textbf{The illustration of the network completion problem}. A network completion problem is to infer the potential links between the unobserved \iffalse vertices\fi nodes and the others according to the observed sub-graph. Left: suppose we can re-arrange all vertices such that the observed nodes come first. Then the entire adjacency matrix $A$ contains two areas: the squared sub-matrix of the observed part $A_o$ and the inverted L-shaped area with the blank entries around $A_o$. Right: the complete adjacency matrix $\hat{A}=A_o\bigoplus \hat{A}_m$, where $\hat{A}_m$ is inferred by the network completion algorithm.}
\label{fig:problem}
\end{figure}

Suppose there is a large undirected network $G(V,E)$ with an adjacency matrix $A$ which can not be fully observed since the information about some nodes and their corresponding edges are missing. In contrast, we are able to observe the sub-graph $G_o$ of $G$, which contains some observed vertices $V_o$ and edges $E_o$ between them. Assume we know the number of missing nodes $N_m$. Our mission is to infer the missing part of the whole network $G_m=G-G_o$ including the missing nodes $V_m$ and the missing edges $E_m$. We can reorder the nodes with the observable nodes in front such that the adjacency matrix $A$ can be decomposed into two regions, one is the sub-matrix of observable nodes ($A_o$) and the other is all the connections related to unobserved nodes ($A_m$). Thus, the task is to reconstruct the whole adjacency matrix $A=A_o\bigoplus A_m$ according to $A_o$, where $A_m$ is an inverted L-shaped matrix which describes the connection between $V_o$ and $V_m$, as well as $V_m$ and $V_m$ as shown in Fig.\ref{fig:problem}, and $\bigoplus$ represents the concatenation operation on adjacency matrices.  

\subsection{Proposed Framework}
\begin{figure}[!ht]
\centering
\includegraphics[scale=0.5]{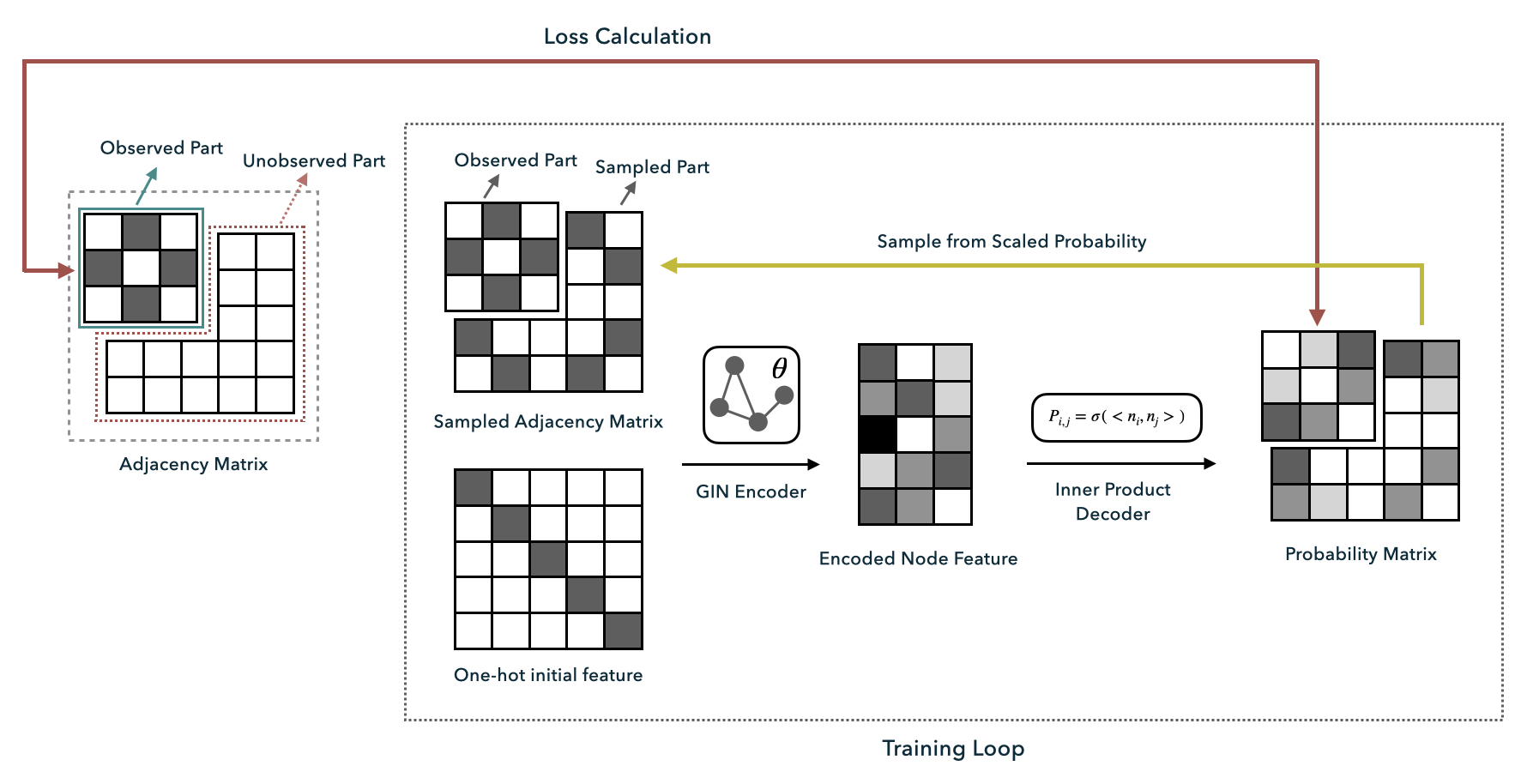}
\caption{\textbf{The Overview of our C-GIN model:} Our model uses GIN to learn the local connection patterns of the known part of the network and complete the missing part in an iterative manner. First, each iteration is separated into encoding stage and decoding stage. In the encoding stage, a set of one-hot vectors which represent the initial node features, and the inferred complete network from the previous iteration are fed into GIN, after that, a set of updated node feature vectors can be obtained as the output. In the second stage, the decoder will generate a matrix representing the probabilities of each node pair according to the updated feature vectors in the previous stage. The observable part of this matrix of probabilities will be used to calculate the loss, and the rest part of the matrix will be re-scaled by multiplying a scaling factor $\gamma$, finally the adjacency matrix as a candidate completion for the next epoch can be sampled according to the probabilities.}
\label{fig:arch}
\end{figure}

The basic assumption behind C-GIN framework is that different parts of the network have similar connection patterns. For example, there is often a local structure of triangles in a social network of acquaintances (one's friends are usually also friends with each other). In C-GIN framework, we will use graph auto-encoder to learn local connection patterns from the partial observed sub-graph and generalize it to the unobserved part. What is ``Learning local connection patterns of networks using Graph Auto-Encoder``? Understand in an intuitive way, learning the local connection pattern of the network is essentially to let the GNN in the Encoder part of the  Graph Auto-Encoder learn how nodes are often linked to form a local structure. This process can be divided into two sub-steps. First, nodes are given initial embedding vectors through a linear neural network layer. Second, through the message passing layers on the network structure, the node and neighbor vectors pass information to each other to form the final embedding vector, thereby encoding the network structure. Many GNN models, such as GCN \cite{kipf2016semi},GAT \cite{2017Graph},GraphSAGE \cite{hamilton2017inductive}, follow this paradigm, but they have different expressive power (the ability to distinguish network structure) due to differences in message passing mechanisms. Since Graph Isomorphism Network(GIN) \cite{xu2018powerful} has been shown to have the strongest expressive power, we use GIN to encode the network structure. But how do we understand "generalize learned pattern to the unobserved part"? We know that Graph Auto-Encoder encodes the network structure into the node embedding vectors, and the node embedding vectors correspond to the network structure. The so-called "generalize learned pattern to the unobserved part" means that by assigning appropriate initial embedding vectors to unobserved nodes, we obtain the complete network structure where unobserved nodes and observed nodes have similar local structures. Fig.\ref{fig:arch} shows the working pipeline of our model.

Next, we will introduce the specific execution flow of the algorithm. According to the problem definition, we have the part information of how the observed nodes connect, namely $A_o$, which is the upper left part of the adjacency matrix.We also assume that we know the number of missing nodes $N_m$. At the first stage, we will fill the adjacency matrix of the unknown part with 0 to get an $N \times N$ matrix $\hat{A}$. We use one-hot vectors as the initial features $X$ of all nodes, so that the linear layers of the neural network can independently determine the embedding vector for each node. Then we feed $\hat{A}$ and $X$ into the GIN encoder, who has been proved to be powerful to learn local connection patterns \cite{xu2018powerful}, to update node feature $H \in \mathbb{R}^{N \times d}$. This process can be expressed by:

\begin{equation}
H = GIN_\theta(\hat{A},X),
\label{eq:eq1}
\end{equation}

\noindent in which $\theta$ represents the parameters of the GIN encoder. We can then feed the encoded node features into the decoder with Eq.\ref{eq:eq2} and get the decoded probability matrix $P_{N \times N}$. $P_{i,j}$ means the probability that node $v_i$ and node $v_j$ may connect:

\begin{equation}
P_{i,j} = \frac{1}{1+\exp(-\langle H^i,H^j \rangle)}.
\label{eq:eq2}
\end{equation}
Note that the probability matrix $P$ is consists of two parts. The upper left part of $P$, i.e., the $N_o \times N_o$ matrix, represents the connection probabilities of the observed part of the network, this part can be used to calculate the loss function and optimize the parameters $\theta$ in the GIN encoder. The loss function is defined by: 

\begin{equation}
\mathcal{L}(\theta) = -\displaystyle\sum_{i=0}^{N_o-2} \sum_{j=i+1}^{N_o} A_{i,j}\log(P_{i,j}) + (1-A_{i,j})\log(1-P_{i,j}).
\label{eq:eq3}
\end{equation}

Besides the upper left region of $P$, the rest of $P$(the inverted L-shaped region) should represent the probabilities of which an unobserved edge may exist for given node pair. However, we cannot sample the inferred adjacency matrix directly according to the unknown part of $P$ because the magnitude of the probabilities in the unobserved part may be too large to keep the edge density consistent with the known part. Therefore, in the second stage of training, we first re-scale the probabilities such that the edge densities of the two parts may be consistent, and then, we can sample edges from the re-scaled matrix. That is, we sample from the re-scaled probability matrix to get the updated adjacency matrix, the rescaling can be done by multiplying a factor $\gamma$ such that $P_{i,j} \times \gamma=1$ for each $i,j$. Thus $\gamma$ can be calculated by:

\begin{equation}
\gamma = \frac{N^2-N_o^2}{N_o^2}  \times  \frac{\sum_{i<N_o,j<N_o}A_{i,j}}{\sum_{i<N,j<N}P_{i,j} - \sum_{i<N_o,j<N_o}P_{i,j}}.
\label{eq:eq6}
\end{equation}

The scaling factor $\gamma$ can control the sparsity of the unobserved part of the predicted adjacency matrix. This is important because the edge density in the unobserved part has a strong influence on the whole candidate network and may cause the graph auto-encoder fail to embed the observed nodes efficiently. To save computing resources, in practice, we don't sample the predicted adjacency matrix in each training epoch, but set the sample interval as a hyper-parameter $\epsilon$. Experiments show that the change of this hyper-parameter has little effect on the final result.

\section{Experiments}

In this section we set up some experiments to demonstrate the capabilities and properties of C-GIN model. First, in subsection \ref{exp1}, we will introduce the baseline models used in different experiments for comparison, they are the Preferential Attachment based method, Random-De method, the G-GCN model and the KronEM model. In subsection \ref{exp2}, we will introduce the metrics we use to measure model performance: Area Under Curve(AUC) and Average Precision Score (AP). Due to the alignment problem, we also introduce the Seeded Graph Matching(SGM) model in this subsection, which is  a new algorithm to align the nodes between learned network and the ground truth, to help measure the performance of network completion. In the network completion task, we will measure the completion performance of edges between observed nodes and unobserved nodes, and the edges between unobserved nodes, so as to better analyze the effect of network completion methods. We will show the network completion performance of our model on synthetic networks in subsection \ref{exp3}, and on real-world networks in subsection \ref{exp4}. Next, we will analyze the properties of our method in subsection \ref{exp5} and propose the 'Reachable CC' indicator. The experiments show that our method tends to perform better when the Reachable CC is higher. We also conduct ablation experiments in the subsection\ref{exp6} to discuss the necessity of scaling factors and alternative calculation schemes. Finally, in subsection\ref{exp7},we compared the completion effect of different GNNs inside the encoders and find that GIN has the best overall performance.

\subsection{Baselines Models}
\label{exp1}
The baselines were chosen from three different types of network completion algorithms for comparison: 

\subsubsection{Preferential Attachment}
Preferential Attachment(PA) is a common baseline in link prediction task. In this model we calculate the score of one node pair, and then we normalize that score to get the probability of the connection. The score of one node pair can be calculated as following equations: 

\begin{equation}
Score(i,j) = D(i)  \times  D(j),
\label{eq:eq4}
\end{equation}

\begin{equation}
P_{i,j} = \frac{Score(i,j)}{\mathop{\max_{i,j}}(Score(i,j))},
\label{eq:eq5}
\end{equation}
\noindent where $D(x)$ means the degree of node $x$ and $P_{i,j}$ means the probability that node $i$ and node $j$ are connected.

We extend the PA model for the network completion problem: if the node pair contains one unobserved node, we only use the degree of the observed nodes as the score (by setting the degree of unobserved node as 1) because the degree of the unobserved node is unavailable. It is reasonable since the unobserved will be more likely to link with the observed node which has more neighbors.


\subsubsection{Random-De}
This method uses the same framework as our proposed approach, except that we abandon the GIN model and instead use randomly generated node embeddings of the same size as the decoder's input. This method can be understood as a kind of random guess. By comparing with this method, we can verify whether our model has learned useful patterns. Besides, we can see that how the presence of GIN module can effect on the model, thus the experiments implemented on this can be also understood as a kind of ablation study.

\subsubsection{G-GCN}
As we explained in the previous section, G-GCN is a model that solves the network completion problem from the perspective of network growth. The original G-GCN model needs to use node feature information as input, this does not fit our definition of the problem, so here we change the input of this model to the one-hot vector, which is the same as our model. Experiments show that the G-GCN model without node feature is still competitive in some tasks.

\subsubsection{KronEM}

KromEM, which is an algorithm of network completion utilizing the self-similarity property of the network, uses the Kronecker graph model to describe the network and estimates the missing part of the network. The Expectation Maximization (EM) framework and a scalable Metropolized Gibbs sampling approach were used to optimize the Kronecker model parameters.

It is important to note that we do not compare this approach on real-world networks, since the number of nodes in the complete graph that must be a power of the size of the kronecker product kernel according to this model which limits its application.

\subsection{Metrics}
\label{exp2}
The completion of the inverted L region of the adjacency matrix can be understood as a task of binary classification, and we evaluate the performances of our model and baselines on the basis of two metrics: the area under the ROC curve (AUC) and the average precision (AP). As previous literature suggests \cite{xu2019generative}, we randomly sampled equal numbers of negative and positive edges when evaluating AUC and AP.

Moreover, we are particularly interested in evaluating the performance over the part where both endpoints of an edge are in the unobserved part. Therefore, we use two sets of metrics, namely $AUC_{Observed - Unobserved}$ and $AP_{Unobserved - Unobserved}$, to evaluate the performances of the connections between observed nodes and unobserved nodes, and the connections between unobserved nodes, respectively.

To calculate AUC and AP, we need to compare the predicted connection probability matrix with a ground truth adjacency matrix, however, we can not do this comparison directly because the inferred unobserved nodes need to be aligned with the actual unobserved nodes. That's to say, for the probability matrix returned by the network completion algorithm, we need a permutation matrix to reorder the rows and columns corresponding to the unobserved nodes so that the probability matrix after the permutation resembles the adjacency matrix as much as possible. There are $N_m$ possible permutations, and for fairness, we choose the best permutation we can find when comparing algorithm performance. This is actually a hard problem which is called sub-graph matching. However, this problem can be solved by the method reported in  \cite{fishkind2019seeded} by optimizing the loss function in:

\begin{equation}
\mathop{\arg\min}\limits_{Q \in \mathcal{Q}} ||A - QPQ^T ||_F^2,
\label{eq:eq7}
\end{equation}

\noindent where $A$ and $P$ represent the ground truth adjacency matrix and the predicted probability matrix, respectively; $Q\in \mathcal{Q}$ is a permutation matrix. $\mathcal{Q}$ is the set of all possible permutations but keeping the sub-matrix of the first $N_o*N_o$ entries corresponding to the observed nodes(the upper left region) being a diagonal matrix. Therefore, $QPQ^T$ means to reorder the rows and columns of the sub-matrix for the unobserved nodes and remain the observable part unchanged. $||\cdot||_F^2$ means the Frobenius norm on matrices. Then the aligning problem can be defined as an optimization problem. The reason why we find a proper $Q$ is because we try to find a node alignment such that each of the unobserved node that we inferred has a local connection pattern to be similar to that of a real unobserved nodes as possible as it can. 


To solve this problem, at the first glimpse, we need to iterate through all possible unobserved permutations and find the best one. However, the number of permutations is $N_m!$, which makes brute force search impossible. Here we introduce the Seeded Graph Matching algorithm \cite{fishkind2019seeded}, which relaxes the problem by allowing the matrix $P$ to be a doubly stochastic matrix whose entities' values range from 0 to 1. Then they use the conjugated optimization method to find an approximate suitable $P$ within a reasonable time complexity ($O(N^3)$). According to their description, the matching accuracy can be more than 90\% when the similarity of two matrices is more than 90\% and the number of nodes to be matched is more than 15. Details can be referred to  \cite{fishkind2019seeded}.


Note that we did not use the Seeded Graph Matching method to reorder the probability matrix returned by KronEM, because this algorithm cannot only output the learned matrix of $A_m$ but also the node alignment.

\subsection{Performances on completing synthetic graphs}
\label{exp3}
In this section we test the performances on completing the synthetic networks which includes 4 types of networks generated by well-known models, namely, Barabási-Albert(BA) model, Watts–Strogatz(WS) model, Forest Fire(FF) model, and Kronecker graphs(Kron) model, respectively. In each network, the number of nodes is 1,024 to satisfy the requirement of KronEM algorithm. In BA network, each new node is added to the network with two connections. In WS network, each node has four neighbors, and the reconnection probability is 0.2. In FF network the probability of both forward and backword of an edge is 0.33. In Kron network, just like the KronEM paper \cite{kim2011network}, we use the $[[0.9,0.7],[,0.5,0.2]]$ as the Kronecker Kernel to generate the network. We randomly choose 25\% of nodes and the relevant edges to remove. Results are shown in Table \ref{tab:tab1}.

\begin{table}[]
\centering
\begin{tabular}{cccccc}
\hline
Networks & PA           &Random-De & KronEM       &  G-GCN              & C-GIN     \\ \hline
BA       & 58.67 ±1.5&59.33 ±1.5 & 64.1 ±0.6 &    74.67 ±2.1    &\textbf{76.49 ±1.8} \\
WS       & 35.45 ±0.5&35.91 ±0.4 & 80.42 ±1.2 &   81.20 ±0.6   &\textbf{85.22 ±0.3} \\
Kron     & 71.60 ±0.6&71.13 ±1.1 & \textbf{83.9 ±0.4} & 68.38 ±1.1 & 71.55 ±1.4 \\
FF       & 79.33 ±0.9&74.16 ±0.7 & 62.11 ±0.4 & \textbf{82.75 ±0.7}  & 80.17±1.8 \\ \hline
\end{tabular}
\caption{\textbf{AUC on unobserved part of synthetic networks:} In this table we show the experimental results of different methods on the synthetic networks.}
\label{tab:tab1}
\end{table}

In Table \ref{tab:tab1} we can see that in most cases, our model performs better than the comparison models. The KronEM model only performs best on the networks generated by the Kron network generator which possesses self-similar property. And G-GCN can get highest score on FF networks which are generated by the forest fire model. In this case, our model also has a competitive performance. Although the BA network is generated by the preference attachment mechanism, the experiments show that the preference attachment method cannot complement the BA network well, because we completely randomly select a part of the nodes to delete, rather than directly delete the last nodes that joined the network. This results in that some large-degree nodes may be deleted, but they will not be connected with a high probability to the small-degree nodes that were once connected to them.


\subsection{Performances on completing real-world graphs}
\label{exp4}
In this section we test the performances of C-GIN model on real-world networks. Different types of real-world networks are selected to test, and their basic information is shown in Table\ref{tab:tab3}.

\begin{itemize}
    \item \textbf{Bio\_S} is an undirected and weighted networks of gene interactions extracted from C. elegans. The nodes are genes and the edges are Inferred links by genetic interactions. To create an unweighted network, we set all the weights greater than zero to be one.
    \item \textbf{Bio\_D}, like Bio\_S, is also an undirected networks of gene interactions extracted from C. elegans. The difference is the density of the network. From Table\ref{tab:tab3}, we can see that Bio\_S is relatively sparse while Bio\_D is relatively dense.
    \item \textbf{Co-Author} network represents the co-authorship of researchers in network theory \& experiments, where a node is a researcher and an edge represents the co-author relationship. If two authors both contribute to at least one paper, a connection between them will be added in the network. The above data is collected from  \cite{nr}.
    \item \textbf{Cora} is a dataset composed of machine learning papers. It is one of the most popular data sets for graph deep learning in recent years, each paper in this dataset has at least one citation. Cora was originally a directed graph. As a conventional operation \cite{kipf2016semi}, we remove all link directions to make it an undirected graph.
\end{itemize}

\begin{table}[]
\centering
\begin{tabular}{@{}ccccc@{}}
\toprule
Network   & Nodes & Edges & Density & Clustering Coefficient \\ \midrule
Bio\_S    & 924   & 3239  & 0.0076  & 0.6051                 \\ \hline
Bio\_D    & 636   & 3959  & 0.0196  & 0.4712                 \\ \hline
Cora      & 2708  & 5278  & 0.0014  & 0.2406                 \\ \hline
Co-Author & 379   & 914   & 0.0128  & 0.7412                 \\ \bottomrule
\end{tabular}
\caption{\textbf{The statistics of different empirical networks:}In this table we show the statistical properties of the networks, they are the number of nodes, the number of edges, the density and the clustering coefficient.}
\label{tab:tab3}
\end{table}

Note that in this experiment we didn't use KronEM as a compared model for the reason that in KronEM model we have to define a kernel and the number of nodes has to be a power of the kernel size, which is not satisfied by most real-world networks. For example, if we define the kernel size to be $2 \times 2$, then the adjacency matrix we want to complete has to be 2 to the $k$ power like 1,024 or 4,096. 

In this experiment, we divided the unobserved part of the adjacency matrix into different regions in order to gain a more detailed understanding of the performance of the model. These regions are 'All', 'Observed-Unobserved' and 'Observed-Unobserved'. 'All' means all the unknown items in the adjacency matrix, 'Observed-Unobserved' means the region describing the connections between observed nodes and unobserved nodes, and 'Unobserved-Unobserved' refers specifically to links between unobserved nodes. Results are shown in Table\ref{tab:tab2}. In all networks, we randomly removed 20\% of the nodes and the corresponding edges, and we ran five repeated experiments to get the mean and the standard deviation to fill the table. To see the results when 10\% and 30\% nodes are removed or the results measured by AP, please refer to the SI for complete information.

\begin{table}[]
\centering
\begin{tabular}{|c|c|cccc|}
\hline
\multirow{2}{*}{Networks}   & \multirow{2}{*}{AUC} & \multicolumn{4}{c|}{Models}                                                                                                       \\ \cline{3-6} 
                           &                          & \multicolumn{1}{c|}{PA}          & \multicolumn{1}{c|}{G-GCN}              & \multicolumn{1}{c|}{C-GIN}            & Random-De \\ \hline
\multirow{3}{*}{Bio\_S}    & All                      & \multicolumn{1}{c|}{72.89 ± 1.3} & \multicolumn{1}{c|}{83.23±1.8}          & \multicolumn{1}{c|}{\textbf{88.71±2.1}}  & 70.22±0.7 \\ \cline{2-6} 
                           & Observed-Unobserved             & \multicolumn{1}{c|}{73.5 ± 1.4}  & \multicolumn{1}{c|}{88.83±1.7}          & \multicolumn{1}{c|}{\textbf{90.16±1.7}}  & 72.41±0.9 \\ \cline{2-6} 
                           & Unobserved-Unobserved              & \multicolumn{1}{c|}{-}            & \multicolumn{1}{c|}{57.19±3.3}          & \multicolumn{1}{c|}{\textbf{74.05±3.4}}  & 54.34±2.2 \\ \hline
\multirow{3}{*}{Bio\_D}    & All                      & \multicolumn{1}{c|}{75.18 ±0.9}  & \multicolumn{1}{c|}{81.35±1.0}          & \multicolumn{1}{c|}{\textbf{85.79±4.0}}  & 62.43±1.6 \\ \cline{2-6} 
                           & Observed-Unobserved              & \multicolumn{1}{c|}{77.01 ± 0.8} & \multicolumn{1}{c|}{85.86±1.0}          & \multicolumn{1}{c|}{\textbf{88.48±2.6}}  & 63.99±1.8 \\ \cline{2-6} 
                           & Unobserved-Unobserved              & \multicolumn{1}{c|}{-}           & \multicolumn{1}{c|}{57.07±1.1}          & \multicolumn{1}{c|}{\textbf{66.71±11.2}} & 54.66±2.7 \\ \hline
\multirow{3}{*}{Cora}      & All                      & \multicolumn{1}{c|}{60.01 ± 0.9} & \multicolumn{1}{c|}{86.02±0.9} & \multicolumn{1}{c|}{\textbf{87.37±0.3}}           & 78.28±1.4 \\ \cline{2-6} 
                           & Observed-Unobserved              & \multicolumn{1}{c|}{60.13 ± 0.6} & \multicolumn{1}{c|}{\textbf{92.07±0.9}} & \multicolumn{1}{c|}{89.10±0.8}           & 81.28±1.1 \\ \cline{2-6} 
                           & Unobserved-Unobserved              & \multicolumn{1}{c|}{-}           & \multicolumn{1}{c|}{55.58±2.3}          & \multicolumn{1}{c|}{\textbf{74.70±2.2}}  & 53.78±2.1 \\ \hline
\multirow{3}{*}{Co-Author} & All                      & \multicolumn{1}{c|}{58.88 ± 2.7} & \multicolumn{1}{c|}{85.82±1.7}          & \multicolumn{1}{c|}{\textbf{91.61±1.6}}  & 73.93±1.5 \\ \cline{2-6} 
                           & Observed Uunobserved              & \multicolumn{1}{c|}{59.07 ± 2.2} & \multicolumn{1}{c|}{97.17±1.1}          & \multicolumn{1}{c|}{\textbf{93.78±1.8}}  & 76.27±1.6 \\ \cline{2-6} 
                           & Unobserved-Unobserved              & \multicolumn{1}{c|}{-}           & \multicolumn{1}{c|}{57.36±7.0}          & \multicolumn{1}{c|}{\textbf{75.29±7.6}}  & 60.66±4.1 \\ \hline
\end{tabular}
\caption{\textbf{AUC on unobserved part of synthetic networks:} In this table we show the experiment result of different methods on synthetic networks.}
\label{tab:tab2}
\end{table}

In Table \ref{tab:tab2} we can see, C-GIN model outperforms all of the competitors in completing the unobserved-unobserved part of the adjacency matrix, which means our model is better at modeling connections between unobserved nodes. In the observed-unobserved part, our model also achieved the best results on two biological networks and co-author networks. However, on the power network and the citation network, G-GCN achieved the best performance, this may be due to the fact that G-GCN is inherently suitable for modeling growing networks, whereas power network and citation networks are typical growing networks. In order to further investigate which kind of network is more suitable for our model to complete, we examine the relationship between the model performance on network completion and the structure features in the next section.

\subsection{Performance vs. Reachable Clustering Coefficient}
\label{exp5}

In the encoding stage of our approach, we use the GIN model to gather neighbor information by message passing. The more times we do message passing, the information of higher ordered neighbors can be aggregated by the model. The aggregated information actually represents the local connection structures of a node which is captured by our model. Therefore, compared with the baselines, our model is able to get better results for networks with more complex local connection structures(for example, the neighbors of the nodes are also connected to each other). Next, we illustrate this point with following experiments.

\begin{figure}[!ht]
\centering
\includegraphics[scale=0.5]{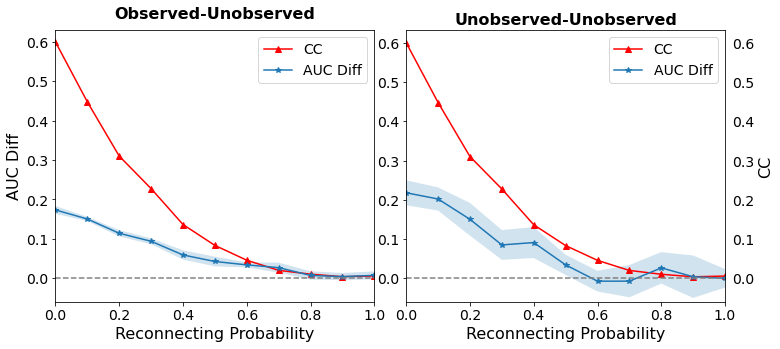}
\caption{\textbf{Performance on W-S network vs. Reconnecting probability}: Indicating the relationship between the AUC(diff) and reconnecting probability on W-S network(blue line). Due to the CC of WS network also decreases as the reconnecting probability increases (the red line) , it shows that the performance of our model is positively correlated with CC}
\label{fig:per_ws}
\end{figure}

In the first experiment, we complete the small-world network, and we can observe the effect that how the reconnecting probability $p$ can influence on the performance of network completion continuously. If $p$ is 0, all nodes are connected to their neighbors, and the neighbors are also connected to each other. In this case, the network has high average clustering coefficient(CC). If $p$ is 1, all nodes are connected by the same number of edges randomly, the neighbors of nodes are hardly connected to each other, and the network has low CC. It is worth noting that the change of $p$ will only change the local connection patterns, but will not affect the network density, network degree distribution and other statistical indicators. We use AUC difference between the C-GIN model and the random baseline model to measure the performance of the model.  Fig \ref{fig:per_ws} shows the relationship between the reconnecting probability $p$ and network completion performance as well as the CC.

In Fig \ref{fig:per_ws}, the x-axis is the reconnecting probability $p$ of the WS network. This figure has two y-axes, one is CC and the other is the AUC difference , in which we replace the output of the GIN encoder with the randomly generated matrix that has the same shape. We can see a clear downward trend. As reconnecting probability $p$ increases, the AUC difference between the C-GIN model and the baseline decreases. The AUC difference in the two regions (edges between observed nodes and unobserved nodes and edges between the unobserved nodes) is reduced to 0 approximately as $p$ increases to 1, which means C-GIN model does nothing better than a random guess in this case. In addition, we can see that CC also decreases as $p$ increases. This clearly shows that our model is more suitable to complete networks with higher CCs.  The reason for this same trend is that for networks with higher CC, the neighbors of the nodes are connected with a higher probability, our model can capture this pattern of local connectivity thus perform better than a random guess. 

But CC is not enough to measure the complexity of the local connection structure. For example, on a grid network in two-dimensional space, any first-order neighbors of a node are not connected to each other, but they will be connected to the node's second-order neighbors, thus forming a complex local connection structure and keep CC constant as 0. In order to measure the interconnections of high-order neighbors of a node, we invented the Reachable CC metric, and found in our experiments that for a network with a higher Reachable CC, our model performs better.

To construct the Reachable CC metric, we need to connect the higher-order neighbors of a network to each other to form a new network, and compute clustering coefficient. The adjacency matrix of the newly constructed network $A_n$ can be obtained byfollowing equations:

\begin{equation}
A^{(n)} = sgn(\sum_{i=1}^n A^i) - \sum_{i=1}^{n-1} A^{(i)},
\label{eq:eq8}
\end{equation}

\begin{equation}
A_n = \sum^n_{i=1} A^{(i)}*(1-\lambda)^{i-1},
\label{eq:eq9}
\end{equation}

\noindent where $A^{(1)} = A$. In Eq \ref{eq:eq8}, we get a matrix $A^{(n)}$ that represents the $n$-order connection between nodes, specifically, if there is a path of length $n$ between node $i$ and node $j$, then $A^{(n)}_{i,j} = 1$, otherwise $A^{(n)}_{i,j} = 0$. In Eq \ref{eq:eq9}, we get $A_n$ by weighted summation of $A^{(i)}$. where $\lambda$ refers to the decay index, which can also be understood as the reaching cost corresponding to the path length. If $\lambda$ is 0, it means that node i can reach node j at no cost. If $\lambda$ is 1, it means that node i cannot reach any second-order and above neighbors, and then $A_n$ degenerates into the original adjacency matrix $A$. We calculate the clustering coefficient on the newly obtained $A_n$, resulting in Reachable CC.

\begin{figure}[!ht]
\centering
\includegraphics[scale=0.5]{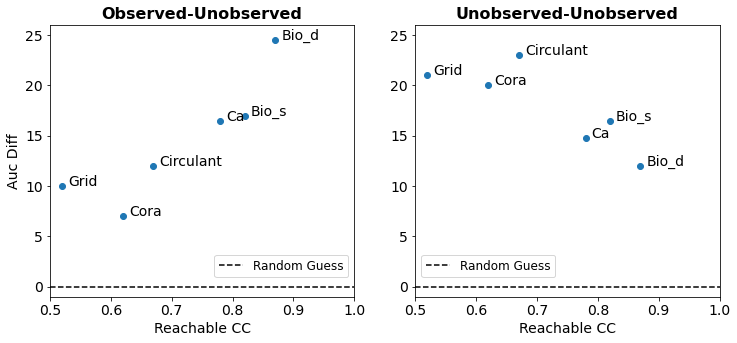}
\caption{\textbf{Performance on Networks vs. Reachable CC}: X-axis is the Reachable CC of networks, y-axis is the performance of network completion. We use the AUC difference between the experimental results and the random guess baseline to compare the results among different networks. The gray dashed line stands for the random guess results.}
\label{fig:rcc_performance}
\end{figure}

In Fig \ref{fig:rcc_performance}, we show how the completion performance of different networks will vary with Reachable CC with $n=2$ and $\lambda=0$, in addition to the four real networks, we also test the Grid network and the Circulant network where each node will connect to the next first and third nodes. We can clearly see that the completion performance of the edge between the observed nodes and unobserved nodes increases with Reachable CC (left panel), and we can also see that the performance of edge completion between unobserved nodes decreases with Reachable CC (right panel). From the working mechanism of the model, we can understand the trend in the left panel to some extent: the connection between the observed nodes are known, so our model can learn the local connection structure pattern of nodes. Due to the setting of the re-scale factor $\gamma$, there must be a certain number of edges between the observed nodes and the unobserved nodes to meet the overall density requirements of the network. Therefore, our model is forced to assign appropriate embedding vectors to unobserved nodes to form edges that conform to the learned pattern. When the Reachable CC is high, the high-order neighbors of nodes are often connected to each other, which is more suitable for the GIN-based encoder to learn the structural pattern of the network. Therefore we can see the completion performance increases with Reachable CC.

\subsection{Ablation Experiment: Does Scaling Factor Necessary?}
\label{exp6}

C-GIN model follows the encoder-decoder architecture and scales the decoded probabilities to sample the completed network. Here, we will demonstrate that this scaling process is necessary by removing or replacing the scaling factor. The purpose of using the scaling factor is to control the density of the sampled adjacency matrix. By scaling the decoded probability matrix and then sampling, the density of the unknown region of the adjacency matrix will be similar to the density of the observed part of the network.

\begin{figure}[!ht]
\centering
\includegraphics[scale=0.45]{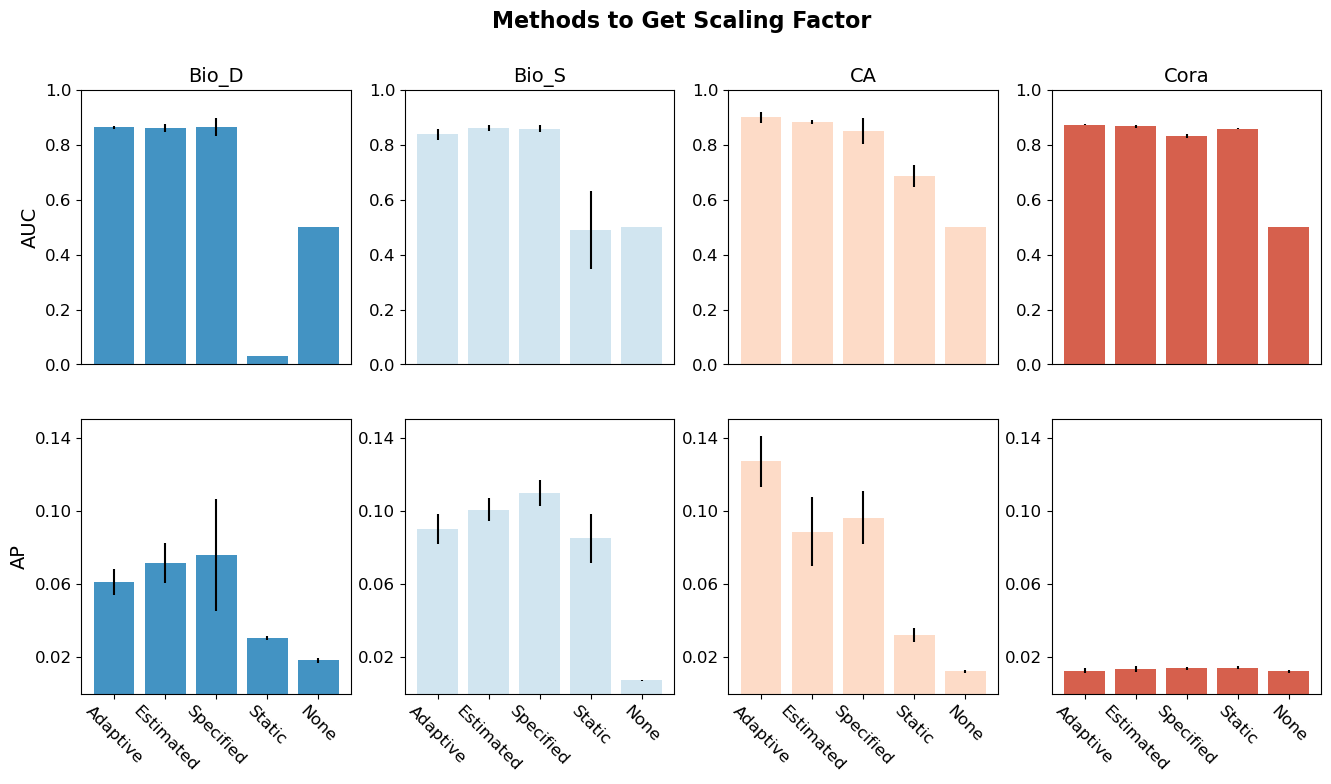}
\caption{\textbf{Candidate methods to calculate the scaling factor}: this figure describes the performance of 5 methods on 4 networks. AUC and AP are used as evaluation metrics. The rows represent AUC and AP respectively, and each column represents a network structure, which have been introduced in section \ref{exp4}}
\label{fig:fig_ablation}
\end{figure}

In the original model, we recalculated the value of the scaling factor every time we sample the predicted adjacency matrix, so throughout the completion process, the value of the scaling factor will continue to change, we call this an 'Adaptive' way. Here we will try to change to other ways to calculate the scaling factor and see if the model still works. Specifically, we will try the following ways and show the results in Fig.\ref{fig:fig_ablation}:

\begin{itemize}
    \item \textbf{None}: We set scaling factor as 1, which means there is no scaling operation.
    \item \textbf{Specified}: We set scaling factor as a specific value(0.001 here).
    \item \textbf{Static}: 
    In this way, we use the decoded probability matrix at the first time in the training process, and calculate the scaling factor by Eq.\ref{eq:eq6}
    
    \item \textbf{Estimated}: In this way, we replace the decoded probability matrix with a uniform random matrix, and then use Eq.\ref{eq:eq6} to calculate the scaling factor.

\end{itemize}

From Fig \ref{fig:fig_ablation} we find that, first of all, the scaling factor is a necessary component, because group 'None' achieves the worst performance in most cases. Second, we also found that the scaling factor is not uniquely calculated, where 'Estimated' and 'Specified' are both possible alternatives, both of which perform even better than the original one on $Bio_D$ and $Bio_S$ networks. Group 'Static' performs worse than the original method in almost all cases, which means that as the training progresses, the probability matrix decoded by the decoder may not be the same as the distribution of the initial case, thus requiring a different scaling factor to keep the density constant .

\subsection{Different GNNs in the Encoder}
\label{exp7}

\begin{figure}[!ht]
\centering
\includegraphics[scale=0.45]{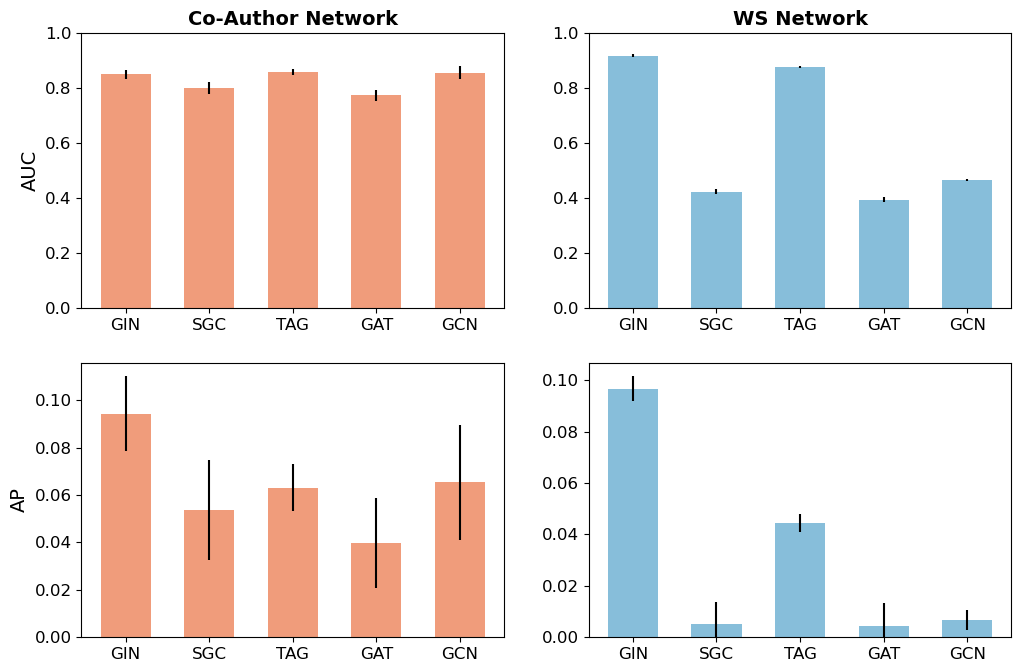}
\caption{\textbf{Performance of Encoder with Different GNNs}:In this figure, we select a typical real network (Co-Author Network, first column) and an synthetic network (WS network, second column) to test different GNNs. We measure the performance of five different GNN models using AUC (first row) and AP (second row) as metrics.}
\label{fig:diff_GNN}
\end{figure}

Our hypothesis is that the model complete the network structure by learning the local connection patterns of the observable network and generalizing. Therefore, the power of Graph Auto-Encoder to express the local connection pattern is crucial. As far as we know, GIN is the most expressive graph network framework. In theory, if we replace GIN with other GNN frameworks with weaker expressive power, the completion performance will be worse. We try to demonstrate this with Fig \ref{fig:diff_GNN}. The comparison GNN models we choose here are:

\begin{itemize}
    \item \textbf{Graph Convolutional Network(GCN)}: GCN \cite{kipf2016semi} is a classic graph neural network structure, and its basic idea is to aggregate neighbor information in a parameterized way. The proposal of GCN has greatly promoted the development of the field of graph networks.
    \item \textbf{Graph Attention Network(GAT)}: GAT \cite{2017Graph} introduces attention into the field of graph networks. When this model aggregates neighbor information, it also calculates the attention weights between neighbors, thereby improving the performance of the model.
    \item \textbf{Topology Adaptive Graph Convolutional Network(TAG)}:TAG \cite{du2017topology} model slides a set of learnable filters to extract both node features and strength of correlation between nodes, thus exhibits better performance than normal GCN.
    \item \textbf{Simplifying Graph Convolutional Networks(SGC)}:SGC \cite{wu2019simplifying} reduced GCN's inherit complexity from deeplearning area by removing nonlinearities and collapsing weight matrices between layers, which gave us a faster, more interpretable GNN model.
\end{itemize}

From Fig \ref{fig:diff_GNN}, we can see that on the real network, most GNNs can achieve good AUC, but only GINs have achieved relatively high AP, which means that every model has a good overall ability to distinguish edges and none-edges, but GIN has the highest accuracy to identify edges. Experiments on WS networks amplify this conclusion, we find that only GIN and TAG can maintain high AUC and AP, while GIN is the best. Therefore, we can prove that GIN has the best overall performance and is most suitable for encoding the network structure.

\section{Conclusion}
In this paper, we propose a model to solve the network completion problem when some nodes and their corresponding edges are missing. This model can learn the local connection patterns of the observable part of the network by Graph Auto-Encode, and generalize them to the unknown region of the adjacency matrix. Experiments show that our model outperforms the comparison models in most synthetic networks and real networks, and is especially good at completing the edges between unobserved nodes. To further investigate the nature of the model, we propose the Reachable CC metric, which essentially measures the probability of connecting edges between higher-order neighbors of nodes in the network. For networks with higher Reachable CC, C-GIN model performs better. Finally, We demonstrate the necessity of individual components in the model through ablation experiments.

Of course, this model still has some limitations. In reality, the number of unobserved nodes is often difficult to obtain, but in this paper, the number of unobserved nodes is assumed to be known. So estimating the number of unobserved nodes would be a meaningful problem. In some cases, we can capture the features of nodes, for example, nodes on Cora's network are papers, which themselves contain feature information. In some cases, there will be some dynamics between nodes, such as SIR dynamics which describing the temporal dynamics of infectious disease. Thus some time series data can be utilized to complete network. How to use the patterns of local structure and the features of these nodes as well as the dynamics together to solve the network completion problem will be the direction of our future study. 

\section*{Acknowledgements}
We are grateful for supporting from Save 2050 Project which is sponsored by Swarma Club and X-Order.
\\
We thank Hao Chen from School of Systems Science, Beijing Normal University, Tiecheng Guo from Physics Department, National University of Singapore and Bingsheng Chen from Imperial College London for discussion and their generous help.

\section*{Funding Statement}
The research is supported by the National Natural Science Foundation of China (NSFC) under the grant numbers 61673070.

\section*{Availability of Code and Data}
The datasets generated and/or analysed during the current study are available in https://github.com/3riccc/C-GIN.

\bibliographystyle{johd}
\bibliography{bib}

\begin{thebibliography}{}

\bibitem [\protect \citeauthoryear {%
Anand%
\ \protect \BOthers {.}}{%
Anand%
\ \protect \BOthers {.}}{%
{\protect \APACyear {2018}}%
}]{%
anand2018missing}
\APACinsertmetastar {%
anand2018missing}%
\begin{APACrefauthors}%
Anand, K.%
, van Lelyveld, I.%
, Banai, {\'A}.%
, Friedrich, S.%
, Garratt, R.%
, Ha{\l}aj, G.%
\BDBL {}others%
\end{APACrefauthors}%
\unskip\
\newblock
\APACrefYearMonthDay{2018}{}{}.
\newblock
{\BBOQ}\APACrefatitle {The missing links: A global study on uncovering
  financial network structures from partial data} {The missing links: A global
  study on uncovering financial network structures from partial data}.{\BBCQ}
\newblock
\APACjournalVolNumPages{Journal of Financial Stability}{35}{}{107--119}.
\PrintBackRefs{\CurrentBib}

\bibitem [\protect \citeauthoryear {%
Betr{\'a}n%
\ \BBA {} Long%
}{%
Betr{\'a}n%
\ \BBA {} Long%
}{%
{\protect \APACyear {2002}}%
}]{%
betran2002expansion}
\APACinsertmetastar {%
betran2002expansion}%
\begin{APACrefauthors}%
Betr{\'a}n, E.%
\BCBT {}\ \BBA {} Long, M.%
\end{APACrefauthors}%
\unskip\
\newblock
\APACrefYearMonthDay{2002}{}{}.
\newblock
{\BBOQ}\APACrefatitle {Expansion of genome coding regions by acquisition of new
  genes} {Expansion of genome coding regions by acquisition of new
  genes}.{\BBCQ}
\newblock
\APACjournalVolNumPages{Genetica}{115}{1}{65--80}.
\PrintBackRefs{\CurrentBib}

\bibitem [\protect \citeauthoryear {%
Bhagat%
, Cormode%
\BCBL {}\ \BBA {} Muthukrishnan%
}{%
Bhagat%
\ \protect \BOthers {.}}{%
{\protect \APACyear {2011}}%
}]{%
bhagat2011node}
\APACinsertmetastar {%
bhagat2011node}%
\begin{APACrefauthors}%
Bhagat, S.%
, Cormode, G.%
\BCBL {}\ \BBA {} Muthukrishnan, S.%
\end{APACrefauthors}%
\unskip\
\newblock
\APACrefYearMonthDay{2011}{}{}.
\newblock
{\BBOQ}\APACrefatitle {Node classification in social networks} {Node
  classification in social networks}.{\BBCQ}
\newblock
\BIn{} \APACrefbtitle {Social network data analytics} {Social network data
  analytics}\ (\BPGS\ 115--148).
\newblock
\APACaddressPublisher{}{Springer}.
\PrintBackRefs{\CurrentBib}

\bibitem [\protect \citeauthoryear {%
Bunke%
\ \BBA {} Riesen%
}{%
Bunke%
\ \BBA {} Riesen%
}{%
{\protect \APACyear {2008}}%
}]{%
bunke2008graph}
\APACinsertmetastar {%
bunke2008graph}%
\begin{APACrefauthors}%
Bunke, H.%
\BCBT {}\ \BBA {} Riesen, K.%
\end{APACrefauthors}%
\unskip\
\newblock
\APACrefYearMonthDay{2008}{}{}.
\newblock
{\BBOQ}\APACrefatitle {Graph classification based on dissimilarity space
  embedding} {Graph classification based on dissimilarity space
  embedding}.{\BBCQ}
\newblock
\BIn{} \APACrefbtitle {Joint IAPR International Workshops on Statistical
  Techniques in Pattern Recognition (SPR) and Structural and Syntactic Pattern
  Recognition (SSPR)} {Joint iapr international workshops on statistical
  techniques in pattern recognition (spr) and structural and syntactic pattern
  recognition (sspr)}\ (\BPGS\ 996--1007).
\PrintBackRefs{\CurrentBib}

\bibitem [\protect \citeauthoryear {%
Cai%
, Zheng%
\BCBL {}\ \BBA {} Chang%
}{%
Cai%
\ \protect \BOthers {.}}{%
{\protect \APACyear {2018}}%
}]{%
cai2018comprehensive}
\APACinsertmetastar {%
cai2018comprehensive}%
\begin{APACrefauthors}%
Cai, H.%
, Zheng, V\BPBI W.%
\BCBL {}\ \BBA {} Chang, K\BPBI C\BHBI C.%
\end{APACrefauthors}%
\unskip\
\newblock
\APACrefYearMonthDay{2018}{}{}.
\newblock
{\BBOQ}\APACrefatitle {A comprehensive survey of graph embedding: Problems,
  techniques, and applications} {A comprehensive survey of graph embedding:
  Problems, techniques, and applications}.{\BBCQ}
\newblock
\APACjournalVolNumPages{IEEE Transactions on Knowledge and Data
  Engineering}{30}{9}{1616--1637}.
\PrintBackRefs{\CurrentBib}

\bibitem [\protect \citeauthoryear {%
Chen%
\ \protect \BOthers {.}}{%
Chen%
\ \protect \BOthers {.}}{%
{\protect \APACyear {2022}}%
}]{%
chen2022inferring}
\APACinsertmetastar {%
chen2022inferring}%
\begin{APACrefauthors}%
Chen, M.%
, Zhang, Y.%
, Zhang, Z.%
, Du, L.%
, Wang, S.%
\BCBL {}\ \BBA {} Zhang, J.%
\end{APACrefauthors}%
\unskip\
\newblock
\APACrefYearMonthDay{2022}{}{}.
\newblock
{\BBOQ}\APACrefatitle {Inferring network structure with unobservable nodes from
  time series data} {Inferring network structure with unobservable nodes from
  time series data}.{\BBCQ}
\newblock
\APACjournalVolNumPages{Chaos: An Interdisciplinary Journal of Nonlinear
  Science}{32}{1}{013126}.
\PrintBackRefs{\CurrentBib}

\bibitem [\protect \citeauthoryear {%
Cimini%
, Squartini%
, Garlaschelli%
\BCBL {}\ \BBA {} Gabrielli%
}{%
Cimini%
\ \protect \BOthers {.}}{%
{\protect \APACyear {2015}}%
}]{%
cimini2015systemic}
\APACinsertmetastar {%
cimini2015systemic}%
\begin{APACrefauthors}%
Cimini, G.%
, Squartini, T.%
, Garlaschelli, D.%
\BCBL {}\ \BBA {} Gabrielli, A.%
\end{APACrefauthors}%
\unskip\
\newblock
\APACrefYearMonthDay{2015}{}{}.
\newblock
{\BBOQ}\APACrefatitle {Systemic risk analysis on reconstructed economic and
  financial networks} {Systemic risk analysis on reconstructed economic and
  financial networks}.{\BBCQ}
\newblock
\APACjournalVolNumPages{Scientific reports}{5}{1}{1--12}.
\PrintBackRefs{\CurrentBib}

\bibitem [\protect \citeauthoryear {%
Do%
, Nguyen%
\BCBL {}\ \BBA {} Deligiannis%
}{%
Do%
\ \protect \BOthers {.}}{%
{\protect \APACyear {2020}}%
}]{%
do2020graph}
\APACinsertmetastar {%
do2020graph}%
\begin{APACrefauthors}%
Do, T\BPBI H.%
, Nguyen, D\BPBI M.%
\BCBL {}\ \BBA {} Deligiannis, N.%
\end{APACrefauthors}%
\unskip\
\newblock
\APACrefYearMonthDay{2020}{}{}.
\newblock
{\BBOQ}\APACrefatitle {Graph auto-encoder for graph signal denoising} {Graph
  auto-encoder for graph signal denoising}.{\BBCQ}
\newblock
\BIn{} \APACrefbtitle {ICASSP 2020-2020 IEEE International Conference on
  Acoustics, Speech and Signal Processing (ICASSP)} {Icassp 2020-2020 ieee
  international conference on acoustics, speech and signal processing
  (icassp)}\ (\BPGS\ 3322--3326).
\PrintBackRefs{\CurrentBib}

\bibitem [\protect \citeauthoryear {%
Du%
, Zhang%
, Wu%
, Moura%
\BCBL {}\ \BBA {} Kar%
}{%
Du%
\ \protect \BOthers {.}}{%
{\protect \APACyear {2017}}%
}]{%
du2017topology}
\APACinsertmetastar {%
du2017topology}%
\begin{APACrefauthors}%
Du, J.%
, Zhang, S.%
, Wu, G.%
, Moura, J\BPBI M.%
\BCBL {}\ \BBA {} Kar, S.%
\end{APACrefauthors}%
\unskip\
\newblock
\APACrefYearMonthDay{2017}{}{}.
\newblock
{\BBOQ}\APACrefatitle {Topology adaptive graph convolutional networks}
  {Topology adaptive graph convolutional networks}.{\BBCQ}
\newblock
\APACjournalVolNumPages{arXiv preprint arXiv:1710.10370}{}{}{}.
\PrintBackRefs{\CurrentBib}

\bibitem [\protect \citeauthoryear {%
Feng%
\ \BBA {} Chen%
}{%
Feng%
\ \BBA {} Chen%
}{%
{\protect \APACyear {2020}}%
}]{%
feng2020link}
\APACinsertmetastar {%
feng2020link}%
\begin{APACrefauthors}%
Feng, J.%
\BCBT {}\ \BBA {} Chen, S.%
\end{APACrefauthors}%
\unskip\
\newblock
\APACrefYearMonthDay{2020}{}{}.
\newblock
{\BBOQ}\APACrefatitle {Link prediction based on orbit counting and graph
  auto-encoder} {Link prediction based on orbit counting and graph
  auto-encoder}.{\BBCQ}
\newblock
\APACjournalVolNumPages{IEEE Access}{8}{}{226773--226783}.
\PrintBackRefs{\CurrentBib}

\bibitem [\protect \citeauthoryear {%
Fishkind%
\ \protect \BOthers {.}}{%
Fishkind%
\ \protect \BOthers {.}}{%
{\protect \APACyear {2019}}%
}]{%
fishkind2019seeded}
\APACinsertmetastar {%
fishkind2019seeded}%
\begin{APACrefauthors}%
Fishkind, D\BPBI E.%
, Adali, S.%
, Patsolic, H\BPBI G.%
, Meng, L.%
, Singh, D.%
, Lyzinski, V.%
\BCBL {}\ \BBA {} Priebe, C\BPBI E.%
\end{APACrefauthors}%
\unskip\
\newblock
\APACrefYearMonthDay{2019}{}{}.
\newblock
{\BBOQ}\APACrefatitle {Seeded graph matching} {Seeded graph matching}.{\BBCQ}
\newblock
\APACjournalVolNumPages{Pattern recognition}{87}{}{203--215}.
\PrintBackRefs{\CurrentBib}

\bibitem [\protect \citeauthoryear {%
Gallos%
, Song%
\BCBL {}\ \protect \BOthers {.}}{%
Gallos%
\ \protect \BOthers {.}}{%
{\protect \APACyear {2007}}%
}]{%
gallos2007review}
\APACinsertmetastar {%
gallos2007review}%
\begin{APACrefauthors}%
Gallos, L\BPBI K.%
, Song, C.%
\BCBL {}\ \BOthersPeriod {.}\end{APACrefauthors}%
\unskip\
\newblock
\APACrefYearMonthDay{2007}{}{}.
\newblock
{\BBOQ}\APACrefatitle {A review of fractality and self-similarity in complex
  networks} {A review of fractality and self-similarity in complex
  networks}.{\BBCQ}
\newblock

\PrintBackRefs{\CurrentBib}

\bibitem [\protect \citeauthoryear {%
Hamilton%
, Ying%
\BCBL {}\ \BBA {} Leskovec%
}{%
Hamilton%
\ \protect \BOthers {.}}{%
{\protect \APACyear {2017}}%
}]{%
hamilton2017inductive}
\APACinsertmetastar {%
hamilton2017inductive}%
\begin{APACrefauthors}%
Hamilton, W.%
, Ying, Z.%
\BCBL {}\ \BBA {} Leskovec, J.%
\end{APACrefauthors}%
\unskip\
\newblock
\APACrefYearMonthDay{2017}{}{}.
\newblock
{\BBOQ}\APACrefatitle {Inductive representation learning on large graphs}
  {Inductive representation learning on large graphs}.{\BBCQ}
\newblock
\APACjournalVolNumPages{Advances in neural information processing
  systems}{30}{}{}.
\PrintBackRefs{\CurrentBib}

\bibitem [\protect \citeauthoryear {%
Kim%
\ \BBA {} Leskovec%
}{%
Kim%
\ \BBA {} Leskovec%
}{%
{\protect \APACyear {2011}}%
}]{%
kim2011network}
\APACinsertmetastar {%
kim2011network}%
\begin{APACrefauthors}%
Kim, M.%
\BCBT {}\ \BBA {} Leskovec, J.%
\end{APACrefauthors}%
\unskip\
\newblock
\APACrefYearMonthDay{2011}{}{}.
\newblock
{\BBOQ}\APACrefatitle {The network completion problem: Inferring missing nodes
  and edges in networks} {The network completion problem: Inferring missing
  nodes and edges in networks}.{\BBCQ}
\newblock
\BIn{} \APACrefbtitle {Proceedings of the 2011 SIAM international conference on
  data mining} {Proceedings of the 2011 siam international conference on data
  mining}\ (\BPGS\ 47--58).
\PrintBackRefs{\CurrentBib}

\bibitem [\protect \citeauthoryear {%
Kipf%
\ \BBA {} Welling%
}{%
Kipf%
\ \BBA {} Welling%
}{%
{\protect \APACyear {2016}}%
{\protect \APACexlab {{\protect \BCnt {1}}}}}]{%
kipf2016semi}
\APACinsertmetastar {%
kipf2016semi}%
\begin{APACrefauthors}%
Kipf, T\BPBI N.%
\BCBT {}\ \BBA {} Welling, M.%
\end{APACrefauthors}%
\unskip\
\newblock
\APACrefYearMonthDay{2016{\protect \BCnt {1}}}{}{}.
\newblock
{\BBOQ}\APACrefatitle {Semi-supervised classification with graph convolutional
  networks} {Semi-supervised classification with graph convolutional
  networks}.{\BBCQ}
\newblock
\APACjournalVolNumPages{arXiv preprint arXiv:1609.02907}{}{}{}.
\PrintBackRefs{\CurrentBib}

\bibitem [\protect \citeauthoryear {%
Kipf%
\ \BBA {} Welling%
}{%
Kipf%
\ \BBA {} Welling%
}{%
{\protect \APACyear {2016}}%
{\protect \APACexlab {{\protect \BCnt {2}}}}}]{%
kipf2016variational}
\APACinsertmetastar {%
kipf2016variational}%
\begin{APACrefauthors}%
Kipf, T\BPBI N.%
\BCBT {}\ \BBA {} Welling, M.%
\end{APACrefauthors}%
\unskip\
\newblock
\APACrefYearMonthDay{2016{\protect \BCnt {2}}}{}{}.
\newblock
{\BBOQ}\APACrefatitle {Variational graph auto-encoders} {Variational graph
  auto-encoders}.{\BBCQ}
\newblock
\APACjournalVolNumPages{arXiv preprint arXiv:1611.07308}{}{}{}.
\PrintBackRefs{\CurrentBib}

\bibitem [\protect \citeauthoryear {%
Kossinets%
}{%
Kossinets%
}{%
{\protect \APACyear {2006}}%
}]{%
kossinets2006effects}
\APACinsertmetastar {%
kossinets2006effects}%
\begin{APACrefauthors}%
Kossinets, G.%
\end{APACrefauthors}%
\unskip\
\newblock
\APACrefYearMonthDay{2006}{}{}.
\newblock
{\BBOQ}\APACrefatitle {Effects of missing data in social networks} {Effects of
  missing data in social networks}.{\BBCQ}
\newblock
\APACjournalVolNumPages{Social networks}{28}{3}{247--268}.
\PrintBackRefs{\CurrentBib}

\bibitem [\protect \citeauthoryear {%
Y.~Li%
, Meng%
, Shahabi%
\BCBL {}\ \BBA {} Liu%
}{%
Y.~Li%
\ \protect \BOthers {.}}{%
{\protect \APACyear {2019}}%
}]{%
li2019structure}
\APACinsertmetastar {%
li2019structure}%
\begin{APACrefauthors}%
Li, Y.%
, Meng, C.%
, Shahabi, C.%
\BCBL {}\ \BBA {} Liu, Y.%
\end{APACrefauthors}%
\unskip\
\newblock
\APACrefYearMonthDay{2019}{}{}.
\newblock
{\BBOQ}\APACrefatitle {Structure-informed graph auto-encoder for relational
  inference and simulation} {Structure-informed graph auto-encoder for
  relational inference and simulation}.{\BBCQ}
\newblock
\BIn{} \APACrefbtitle {ICML Workshop on Learning and Reasoning with
  Graph-Structured Data} {Icml workshop on learning and reasoning with
  graph-structured data}\ (\BVOL~8, \BPG~2).
\PrintBackRefs{\CurrentBib}

\bibitem [\protect \citeauthoryear {%
Z.~Li%
, Li%
, Nie%
, You%
\BCBL {}\ \BBA {} Bao%
}{%
Z.~Li%
\ \protect \BOthers {.}}{%
{\protect \APACyear {2021}}%
}]{%
li2021graph}
\APACinsertmetastar {%
li2021graph}%
\begin{APACrefauthors}%
Li, Z.%
, Li, J.%
, Nie, R.%
, You, Z\BHBI H.%
\BCBL {}\ \BBA {} Bao, W.%
\end{APACrefauthors}%
\unskip\
\newblock
\APACrefYearMonthDay{2021}{}{}.
\newblock
{\BBOQ}\APACrefatitle {A graph auto-encoder model for miRNA-disease
  associations prediction} {A graph auto-encoder model for mirna-disease
  associations prediction}.{\BBCQ}
\newblock
\APACjournalVolNumPages{Briefings in Bioinformatics}{22}{4}{}.
\PrintBackRefs{\CurrentBib}

\bibitem [\protect \citeauthoryear {%
Lichtenwalter%
, Lussier%
\BCBL {}\ \BBA {} Chawla%
}{%
Lichtenwalter%
\ \protect \BOthers {.}}{%
{\protect \APACyear {2010}}%
}]{%
lichtenwalter2010new}
\APACinsertmetastar {%
lichtenwalter2010new}%
\begin{APACrefauthors}%
Lichtenwalter, R\BPBI N.%
, Lussier, J\BPBI T.%
\BCBL {}\ \BBA {} Chawla, N\BPBI V.%
\end{APACrefauthors}%
\unskip\
\newblock
\APACrefYearMonthDay{2010}{}{}.
\newblock
{\BBOQ}\APACrefatitle {New perspectives and methods in link prediction} {New
  perspectives and methods in link prediction}.{\BBCQ}
\newblock
\BIn{} \APACrefbtitle {Proceedings of the 16th ACM SIGKDD international
  conference on Knowledge discovery and data mining} {Proceedings of the 16th
  acm sigkdd international conference on knowledge discovery and data mining}\
  (\BPGS\ 243--252).
\PrintBackRefs{\CurrentBib}

\bibitem [\protect \citeauthoryear {%
Liu%
, Zhang%
, L{\"u}%
\BCBL {}\ \BBA {} Zhou%
}{%
Liu%
\ \protect \BOthers {.}}{%
{\protect \APACyear {2011}}%
}]{%
liu2011link}
\APACinsertmetastar {%
liu2011link}%
\begin{APACrefauthors}%
Liu, Z.%
, Zhang, Q\BHBI M.%
, L{\"u}, L.%
\BCBL {}\ \BBA {} Zhou, T.%
\end{APACrefauthors}%
\unskip\
\newblock
\APACrefYearMonthDay{2011}{}{}.
\newblock
{\BBOQ}\APACrefatitle {Link prediction in complex networks: A local na{\"\i}ve
  Bayes model} {Link prediction in complex networks: A local na{\"\i}ve bayes
  model}.{\BBCQ}
\newblock
\APACjournalVolNumPages{EPL (Europhysics Letters)}{96}{4}{48007}.
\PrintBackRefs{\CurrentBib}

\bibitem [\protect \citeauthoryear {%
Long%
, Betr{\'a}n%
, Thornton%
\BCBL {}\ \BBA {} Wang%
}{%
Long%
\ \protect \BOthers {.}}{%
{\protect \APACyear {2003}}%
}]{%
long2003origin}
\APACinsertmetastar {%
long2003origin}%
\begin{APACrefauthors}%
Long, M.%
, Betr{\'a}n, E.%
, Thornton, K.%
\BCBL {}\ \BBA {} Wang, W.%
\end{APACrefauthors}%
\unskip\
\newblock
\APACrefYearMonthDay{2003}{}{}.
\newblock
{\BBOQ}\APACrefatitle {The origin of new genes: glimpses from the young and
  old} {The origin of new genes: glimpses from the young and old}.{\BBCQ}
\newblock
\APACjournalVolNumPages{Nature Reviews Genetics}{4}{11}{865--875}.
\PrintBackRefs{\CurrentBib}

\bibitem [\protect \citeauthoryear {%
L{\"u}%
\ \BBA {} Zhou%
}{%
L{\"u}%
\ \BBA {} Zhou%
}{%
{\protect \APACyear {2011}}%
}]{%
lu2011link}
\APACinsertmetastar {%
lu2011link}%
\begin{APACrefauthors}%
L{\"u}, L.%
\BCBT {}\ \BBA {} Zhou, T.%
\end{APACrefauthors}%
\unskip\
\newblock
\APACrefYearMonthDay{2011}{}{}.
\newblock
{\BBOQ}\APACrefatitle {Link prediction in complex networks: A survey} {Link
  prediction in complex networks: A survey}.{\BBCQ}
\newblock
\APACjournalVolNumPages{Physica A: statistical mechanics and its
  applications}{390}{6}{1150--1170}.
\PrintBackRefs{\CurrentBib}

\bibitem [\protect \citeauthoryear {%
Marsden%
}{%
Marsden%
}{%
{\protect \APACyear {1990}}%
}]{%
marsden1990network}
\APACinsertmetastar {%
marsden1990network}%
\begin{APACrefauthors}%
Marsden, P\BPBI V.%
\end{APACrefauthors}%
\unskip\
\newblock
\APACrefYearMonthDay{1990}{}{}.
\newblock
{\BBOQ}\APACrefatitle {Network data and measurement} {Network data and
  measurement}.{\BBCQ}
\newblock
\APACjournalVolNumPages{Annual review of sociology}{16}{1}{435--463}.
\PrintBackRefs{\CurrentBib}

\bibitem [\protect \citeauthoryear {%
Newman%
}{%
Newman%
}{%
{\protect \APACyear {2012}}%
}]{%
newman2012communities}
\APACinsertmetastar {%
newman2012communities}%
\begin{APACrefauthors}%
Newman, M\BPBI E.%
\end{APACrefauthors}%
\unskip\
\newblock
\APACrefYearMonthDay{2012}{}{}.
\newblock
{\BBOQ}\APACrefatitle {Communities, modules and large-scale structure in
  networks} {Communities, modules and large-scale structure in
  networks}.{\BBCQ}
\newblock
\APACjournalVolNumPages{Nature physics}{8}{1}{25--31}.
\PrintBackRefs{\CurrentBib}

\bibitem [\protect \citeauthoryear {%
Pearl%
}{%
Pearl%
}{%
{\protect \APACyear {2010}}%
}]{%
pearl2010causal}
\APACinsertmetastar {%
pearl2010causal}%
\begin{APACrefauthors}%
Pearl, J.%
\end{APACrefauthors}%
\unskip\
\newblock
\APACrefYearMonthDay{2010}{}{}.
\newblock
{\BBOQ}\APACrefatitle {Causal inference} {Causal inference}.{\BBCQ}
\newblock
\APACjournalVolNumPages{Causality: objectives and assessment}{}{}{39--58}.
\PrintBackRefs{\CurrentBib}

\bibitem [\protect \citeauthoryear {%
Rong%
, Huang%
, Xu%
\BCBL {}\ \BBA {} Huang%
}{%
Rong%
\ \protect \BOthers {.}}{%
{\protect \APACyear {2019}}%
}]{%
rong2019dropedge}
\APACinsertmetastar {%
rong2019dropedge}%
\begin{APACrefauthors}%
Rong, Y.%
, Huang, W.%
, Xu, T.%
\BCBL {}\ \BBA {} Huang, J.%
\end{APACrefauthors}%
\unskip\
\newblock
\APACrefYearMonthDay{2019}{}{}.
\newblock
{\BBOQ}\APACrefatitle {Dropedge: Towards deep graph convolutional networks on
  node classification} {Dropedge: Towards deep graph convolutional networks on
  node classification}.{\BBCQ}
\newblock
\APACjournalVolNumPages{arXiv preprint arXiv:1907.10903}{}{}{}.
\PrintBackRefs{\CurrentBib}

\bibitem [\protect \citeauthoryear {%
Rossi%
\ \BBA {} Ahmed%
}{%
Rossi%
\ \BBA {} Ahmed%
}{%
{\protect \APACyear {2015}}%
}]{%
nr}
\APACinsertmetastar {%
nr}%
\begin{APACrefauthors}%
Rossi, R\BPBI A.%
\BCBT {}\ \BBA {} Ahmed, N\BPBI K.%
\end{APACrefauthors}%
\unskip\
\newblock
\APACrefYearMonthDay{2015}{}{}.
\newblock
{\BBOQ}\APACrefatitle {The Network Data Repository with Interactive Graph
  Analytics and Visualization} {The network data repository with interactive
  graph analytics and visualization}.{\BBCQ}
\newblock
\BIn{} \APACrefbtitle {AAAI.} {Aaai.}
\newblock
\begin{APACrefURL} \url{https://networkrepository.com} \end{APACrefURL}
\PrintBackRefs{\CurrentBib}

\bibitem [\protect \citeauthoryear {%
Sanchez-Gonzalez%
\ \protect \BOthers {.}}{%
Sanchez-Gonzalez%
\ \protect \BOthers {.}}{%
{\protect \APACyear {2020}}%
}]{%
sanchez2020learning}
\APACinsertmetastar {%
sanchez2020learning}%
\begin{APACrefauthors}%
Sanchez-Gonzalez, A.%
, Godwin, J.%
, Pfaff, T.%
, Ying, R.%
, Leskovec, J.%
\BCBL {}\ \BBA {} Battaglia, P.%
\end{APACrefauthors}%
\unskip\
\newblock
\APACrefYearMonthDay{2020}{}{}.
\newblock
{\BBOQ}\APACrefatitle {Learning to simulate complex physics with graph
  networks} {Learning to simulate complex physics with graph networks}.{\BBCQ}
\newblock
\BIn{} \APACrefbtitle {International Conference on Machine Learning}
  {International conference on machine learning}\ (\BPGS\ 8459--8468).
\PrintBackRefs{\CurrentBib}

\bibitem [\protect \citeauthoryear {%
Song%
, Havlin%
\BCBL {}\ \BBA {} Makse%
}{%
Song%
\ \protect \BOthers {.}}{%
{\protect \APACyear {2005}}%
}]{%
song2005self}
\APACinsertmetastar {%
song2005self}%
\begin{APACrefauthors}%
Song, C.%
, Havlin, S.%
\BCBL {}\ \BBA {} Makse, H\BPBI A.%
\end{APACrefauthors}%
\unskip\
\newblock
\APACrefYearMonthDay{2005}{}{}.
\newblock
{\BBOQ}\APACrefatitle {Self-similarity of complex networks} {Self-similarity of
  complex networks}.{\BBCQ}
\newblock
\APACjournalVolNumPages{Nature}{433}{7024}{392--395}.
\PrintBackRefs{\CurrentBib}

\bibitem [\protect \citeauthoryear {%
Tan%
, Xia%
\BCBL {}\ \BBA {} Zhu%
}{%
Tan%
\ \protect \BOthers {.}}{%
{\protect \APACyear {2014}}%
}]{%
tan2014link}
\APACinsertmetastar {%
tan2014link}%
\begin{APACrefauthors}%
Tan, F.%
, Xia, Y.%
\BCBL {}\ \BBA {} Zhu, B.%
\end{APACrefauthors}%
\unskip\
\newblock
\APACrefYearMonthDay{2014}{}{}.
\newblock
{\BBOQ}\APACrefatitle {Link prediction in complex networks: a mutual
  information perspective} {Link prediction in complex networks: a mutual
  information perspective}.{\BBCQ}
\newblock
\APACjournalVolNumPages{PloS one}{9}{9}{e107056}.
\PrintBackRefs{\CurrentBib}

\bibitem [\protect \citeauthoryear {%
Tran%
, Shin%
, Spitz%
\BCBL {}\ \BBA {} Gertz%
}{%
Tran%
\ \protect \BOthers {.}}{%
{\protect \APACyear {2020}}%
}]{%
tran2020deepnc}
\APACinsertmetastar {%
tran2020deepnc}%
\begin{APACrefauthors}%
Tran, C.%
, Shin, W\BHBI Y.%
, Spitz, A.%
\BCBL {}\ \BBA {} Gertz, M.%
\end{APACrefauthors}%
\unskip\
\newblock
\APACrefYearMonthDay{2020}{}{}.
\newblock
{\BBOQ}\APACrefatitle {DeepNC: Deep generative network completion} {Deepnc:
  Deep generative network completion}.{\BBCQ}
\newblock
\APACjournalVolNumPages{IEEE Transactions on Pattern Analysis and Machine
  Intelligence}{}{}{}.
\PrintBackRefs{\CurrentBib}

\bibitem [\protect \citeauthoryear {%
Velikovi%
\ \protect \BOthers {.}}{%
Velikovi%
\ \protect \BOthers {.}}{%
{\protect \APACyear {2017}}%
}]{%
2017Graph}
\APACinsertmetastar {%
2017Graph}%
\begin{APACrefauthors}%
Velikovi, P.%
, Cucurull, G.%
, Casanova, A.%
, Romero, A.%
, Liò, P.%
\BCBL {}\ \BBA {} Bengio, Y.%
\end{APACrefauthors}%
\unskip\
\newblock
\APACrefYearMonthDay{2017}{}{}.
\newblock
{\BBOQ}\APACrefatitle {Graph Attention Networks} {Graph attention
  networks}.{\BBCQ}
\newblock

\PrintBackRefs{\CurrentBib}

\bibitem [\protect \citeauthoryear {%
P.~Wang%
, Xu%
, Wu%
\BCBL {}\ \BBA {} Zhou%
}{%
P.~Wang%
\ \protect \BOthers {.}}{%
{\protect \APACyear {2015}}%
}]{%
wang2015link}
\APACinsertmetastar {%
wang2015link}%
\begin{APACrefauthors}%
Wang, P.%
, Xu, B.%
, Wu, Y.%
\BCBL {}\ \BBA {} Zhou, X.%
\end{APACrefauthors}%
\unskip\
\newblock
\APACrefYearMonthDay{2015}{}{}.
\newblock
{\BBOQ}\APACrefatitle {Link prediction in social networks: the
  state-of-the-art} {Link prediction in social networks: the
  state-of-the-art}.{\BBCQ}
\newblock
\APACjournalVolNumPages{Science China Information Sciences}{58}{1}{1--38}.
\PrintBackRefs{\CurrentBib}

\bibitem [\protect \citeauthoryear {%
S.~Wang%
\ \protect \BOthers {.}}{%
S.~Wang%
\ \protect \BOthers {.}}{%
{\protect \APACyear {2020}}%
}]{%
wang2020pm2}
\APACinsertmetastar {%
wang2020pm2}%
\begin{APACrefauthors}%
Wang, S.%
, Li, Y.%
, Zhang, J.%
, Meng, Q.%
, Meng, L.%
\BCBL {}\ \BBA {} Gao, F.%
\end{APACrefauthors}%
\unskip\
\newblock
\APACrefYearMonthDay{2020}{}{}.
\newblock
{\BBOQ}\APACrefatitle {Pm2. 5-gnn: A domain knowledge enhanced graph neural
  network for pm2. 5 forecasting} {Pm2. 5-gnn: A domain knowledge enhanced
  graph neural network for pm2. 5 forecasting}.{\BBCQ}
\newblock
\BIn{} \APACrefbtitle {Proceedings of the 28th International Conference on
  Advances in Geographic Information Systems} {Proceedings of the 28th
  international conference on advances in geographic information systems}\
  (\BPGS\ 163--166).
\PrintBackRefs{\CurrentBib}

\bibitem [\protect \citeauthoryear {%
Y.~Wang%
, Yao%
\BCBL {}\ \BBA {} Zhao%
}{%
Y.~Wang%
\ \protect \BOthers {.}}{%
{\protect \APACyear {2016}}%
}]{%
wang2016auto}
\APACinsertmetastar {%
wang2016auto}%
\begin{APACrefauthors}%
Wang, Y.%
, Yao, H.%
\BCBL {}\ \BBA {} Zhao, S.%
\end{APACrefauthors}%
\unskip\
\newblock
\APACrefYearMonthDay{2016}{}{}.
\newblock
{\BBOQ}\APACrefatitle {Auto-encoder based dimensionality reduction}
  {Auto-encoder based dimensionality reduction}.{\BBCQ}
\newblock
\APACjournalVolNumPages{Neurocomputing}{184}{}{232--242}.
\PrintBackRefs{\CurrentBib}

\bibitem [\protect \citeauthoryear {%
Wei%
\ \BBA {} Hu%
}{%
Wei%
\ \BBA {} Hu%
}{%
{\protect \APACyear {2021}}%
}]{%
wei2021unifying}
\APACinsertmetastar {%
wei2021unifying}%
\begin{APACrefauthors}%
Wei, Q.%
\BCBT {}\ \BBA {} Hu, G.%
\end{APACrefauthors}%
\unskip\
\newblock
\APACrefYearMonthDay{2021}{}{}.
\newblock
{\BBOQ}\APACrefatitle {Unifying Node Labels, Features, and Distances for Deep
  Network Completion} {Unifying node labels, features, and distances for deep
  network completion}.{\BBCQ}
\newblock
\APACjournalVolNumPages{Entropy}{23}{6}{771}.
\PrintBackRefs{\CurrentBib}

\bibitem [\protect \citeauthoryear {%
Wu%
\ \protect \BOthers {.}}{%
Wu%
\ \protect \BOthers {.}}{%
{\protect \APACyear {2019}}%
}]{%
wu2019simplifying}
\APACinsertmetastar {%
wu2019simplifying}%
\begin{APACrefauthors}%
Wu, F.%
, Souza, A.%
, Zhang, T.%
, Fifty, C.%
, Yu, T.%
\BCBL {}\ \BBA {} Weinberger, K.%
\end{APACrefauthors}%
\unskip\
\newblock
\APACrefYearMonthDay{2019}{}{}.
\newblock
{\BBOQ}\APACrefatitle {Simplifying graph convolutional networks} {Simplifying
  graph convolutional networks}.{\BBCQ}
\newblock
\BIn{} \APACrefbtitle {International conference on machine learning}
  {International conference on machine learning}\ (\BPGS\ 6861--6871).
\PrintBackRefs{\CurrentBib}

\bibitem [\protect \citeauthoryear {%
D.~Xu%
\ \protect \BOthers {.}}{%
D.~Xu%
\ \protect \BOthers {.}}{%
{\protect \APACyear {2019}}%
}]{%
xu2019generative}
\APACinsertmetastar {%
xu2019generative}%
\begin{APACrefauthors}%
Xu, D.%
, Ruan, C.%
, Motwani, K.%
, Korpeoglu, E.%
, Kumar, S.%
\BCBL {}\ \BBA {} Achan, K.%
\end{APACrefauthors}%
\unskip\
\newblock
\APACrefYearMonthDay{2019}{}{}.
\newblock
{\BBOQ}\APACrefatitle {Generative graph convolutional network for growing
  graphs} {Generative graph convolutional network for growing graphs}.{\BBCQ}
\newblock
\BIn{} \APACrefbtitle {ICASSP 2019-2019 IEEE International Conference on
  Acoustics, Speech and Signal Processing (ICASSP)} {Icassp 2019-2019 ieee
  international conference on acoustics, speech and signal processing
  (icassp)}\ (\BPGS\ 3167--3171).
\PrintBackRefs{\CurrentBib}

\bibitem [\protect \citeauthoryear {%
K.~Xu%
, Hu%
, Leskovec%
\BCBL {}\ \BBA {} Jegelka%
}{%
K.~Xu%
\ \protect \BOthers {.}}{%
{\protect \APACyear {2018}}%
}]{%
xu2018powerful}
\APACinsertmetastar {%
xu2018powerful}%
\begin{APACrefauthors}%
Xu, K.%
, Hu, W.%
, Leskovec, J.%
\BCBL {}\ \BBA {} Jegelka, S.%
\end{APACrefauthors}%
\unskip\
\newblock
\APACrefYearMonthDay{2018}{}{}.
\newblock
{\BBOQ}\APACrefatitle {How powerful are graph neural networks?} {How powerful
  are graph neural networks?}{\BBCQ}
\newblock
\APACjournalVolNumPages{arXiv preprint arXiv:1810.00826}{}{}{}.
\PrintBackRefs{\CurrentBib}

\bibitem [\protect \citeauthoryear {%
Zhang%
\ \BBA {} Chen%
}{%
Zhang%
\ \BBA {} Chen%
}{%
{\protect \APACyear {2018}}%
}]{%
zhang2018link}
\APACinsertmetastar {%
zhang2018link}%
\begin{APACrefauthors}%
Zhang, M.%
\BCBT {}\ \BBA {} Chen, Y.%
\end{APACrefauthors}%
\unskip\
\newblock
\APACrefYearMonthDay{2018}{}{}.
\newblock
{\BBOQ}\APACrefatitle {Link prediction based on graph neural networks} {Link
  prediction based on graph neural networks}.{\BBCQ}
\newblock
\APACjournalVolNumPages{Advances in neural information processing
  systems}{31}{}{}.
\PrintBackRefs{\CurrentBib}

\bibitem [\protect \citeauthoryear {%
Zhou%
, L{\"u}%
\BCBL {}\ \BBA {} Zhang%
}{%
Zhou%
\ \protect \BOthers {.}}{%
{\protect \APACyear {2009}}%
}]{%
zhou2009predicting}
\APACinsertmetastar {%
zhou2009predicting}%
\begin{APACrefauthors}%
Zhou, T.%
, L{\"u}, L.%
\BCBL {}\ \BBA {} Zhang, Y\BHBI C.%
\end{APACrefauthors}%
\unskip\
\newblock
\APACrefYearMonthDay{2009}{}{}.
\newblock
{\BBOQ}\APACrefatitle {Predicting missing links via local information}
  {Predicting missing links via local information}.{\BBCQ}
\newblock
\APACjournalVolNumPages{The European Physical Journal B}{71}{4}{623--630}.
\PrintBackRefs{\CurrentBib}

\end{thebibliography}



\end{document}